%% file: 0-Main.tex
\documentclass[conference]{IEEEtran}
\IEEEoverridecommandlockouts
\usepackage{cite}
\usepackage{amsmath,amssymb,amsfonts}
\usepackage{algorithmic}
\usepackage{graphicx}
\usepackage{textcomp}
\usepackage{xcolor}
\usepackage{bm}
\usepackage{booktabs}
\usepackage{algorithm}
\usepackage{enumitem}
\usepackage{multirow}
\usepackage{diagbox}
\usepackage{color}
\usepackage{subfigure}
\usepackage{booktabs}
\usepackage{caption}
\usepackage{url}
\usepackage{hyperref}
\usepackage{xurl}

\def\BibTeX{{\rm B\kern-.05em{\sc i\kern-.025em b}\kern-.08em
    T\kern-.1667em\lower.7ex\hbox{E}\kern-.125emX}}
\begin{document}

\title{An Embarrassingly Simple Rule-based Visiting Circulation Approach to Trip Destination Prediction}

\makeatletter
\newcommand{\linebreakand}{%
  \end{@IEEEauthorhalign}
  \hfill\mbox{}\par
  \mbox{}\hfill\begin{@IEEEauthorhalign}
}
\makeatother

\author{
\IEEEauthorblockN{Eng-Shen Tu\IEEEauthorrefmark{1}}
\IEEEauthorblockA{
\textit{National Cheng Kung University}\\
Tainan, Taiwan \\
jamestu6301@gmail.com}
\and
\IEEEauthorblockN{Yong-Han Chen\IEEEauthorrefmark{1}}
\IEEEcompsocitemizethanks{\IEEEcompsocthanksitem\IEEEauthorrefmark{1} Equal contribution to this work.}
\IEEEauthorblockA{
\textit{National Cheng Kung University}\\
Tainan, Taiwan \\
hank96015@gmail.com}
\and
\IEEEauthorblockN{En-Chao Liu}
\IEEEauthorblockA{
\textit{National Cheng Kung University}\\
Tainan, Taiwan \\
danny010324@gmail.com}
\linebreakand
\IEEEauthorblockN{Hao-Yun Keng}
\IEEEauthorblockA{
\textit{National Cheng Kung University}\\
Tainan, Taiwan \\
darinkeng@gmail.com}
\and
\IEEEauthorblockN{Cheng-Te Li}
\IEEEauthorblockA{
\textit{National Cheng Kung University}\\
Tainan, Taiwan \\
chengte@ncku.edu.tw}
}

\maketitle

\begin{abstract}
In this paper, we propose the Rule-based Visiting Circulation (RVC) model in tackling the challenge in the IEEE Big Data Cup 2022: Trip Destination Prediction. Given trips containing travel information, personal attributes, origin zones, and their features in the training metropolitan areas, the task is to predict the destination of every trip in a targeted metropolitan area whose destinations are not given at all at the training stage. We highlight the challenges in this destination prediction task -- having no knowledge of the destinations in the targeted metropolitan area. We provide insights from the datasets, in which revisiting behaviors and the relationships between origins and destinations play a crucial role in individuals' trips. Hence, we design a simple but comprehensive method, rule-based visiting circulation, which directly utilizes the origin information and individuals' trip behaviors to determine the destinations in the targeted metropolitan area, i.e., requiring no learning from the four training areas. Experimental results on both offline evaluation and leaderboard submission consistently exhibit the proposed RVC can significantly outperform supervised learning methods and other heuristics. The RVC method eventually brings us to second place in the competition leaderboard.
\end{abstract}


\input{1-Introduction.tex}

\input{4-ExplorativeAnalysis.tex}

\input{5-Model.tex}
\input{6-Experiments.tex}
\input{7-Conclusion.tex}

\section*{Acknowledgment}
This work is supported by the National Science and Technology Council (NSTC) of Taiwan under grants 110-2221-E-006-136-MY3, 111-2221-E-006-001, and 111-2634-F-002-022. 






\bibliographystyle{plain}
\bibliography{9-Refs-short}


\end{document}

%% file: 1-Introduction.tex
\section{Introduction}
\label{sec-intro}
The IEEE BigData Cup 2022: Trip Destination Prediction\footnote{\url{https://www.kaggle.com/competitions/ieee-bigdata-cup-2022-destination-prediction/}} competition is part of the IEEE BigData Cup Challenge at the IEEE International Conference on Big Data 2022. The overarching aim of the competition is for us to develop a robust and generalizable model to predict an individual's daily trip destination based on easily collected information and features using data collected from multiple metropolitan areas in Japan, namely Chukyo, Higashisurugawan (Higa.), Kyushu, and Tokyo, then further utilize the model to predict trip destinations for individuals in a new metropolitan area, Kinki. The submission files only contain two columns, Tripid and Destination, where Destinations should be one of the ZONE\_IDs found in the corresponding file. Submissions are evaluated based on the Categorization Accuracy of the predicted destinations. As mentioned in the IEEE Big Data Cup 2022: Trip Destination Prediction challenge description, destination prediction not only plays a key role in city governance, emergency regulation, and location-based services, but also helps support policymaking, travel demand estimation, transit planning, and many more applications in our society.

While the goal of the competition is to predict trip destinations of individuals at a citywide level, the predictions made will be mainly based on an individual's personal attributes (i.e., age, gender, and occupation) and trip details (i.e., trip type, departure time, and origin), which are given in the dataset. The baseline of the competition is set where the predicted destination equals the origin of the trip. The accuracy of the baseline is $0.20879$, as indicated by the competition.

The main challenge of this trip destination prediction task is three-fold, as listed below.
\begin{itemize}
\item \textit{Differences between datasets of metropolitan areas}: while there are four given datasets in total, the zone IDs, occupation, and trip type features vary from one dataset to the other, making categorization and labeling difficult for a prediction model to be precisely learned and constructed. Such a severe situation occurs especially when the origins and destinations are completely different for each metropolitan area.

\item \textit{Organizing geographic information}: since the prediction goal of the trip destination is related to destinations and each zone's geographical features, such as the distance could be crucial to making accurate predictions. Although data on geographical features are given (i.e., zone features), going through and dealing with such kind of data is complex and extremely hard to organize.

\item \textit{A large number of destination targets}: with more than $1,000$ possible destinations, it is extremely difficult for a supervised model to learn and generalize the given data and then make accurate destination predictions by corresponding it to a particular zone that is never seen at the model training stage. 
\end{itemize}

With the challenges mentioned above, our proposed solution is a \textit{Rule-based Visiting Circulation} (RVC) model that attempts to overcome these challenges, yields can produce promising prediction results, and proves to be robust towards datasets across different metropolitan areas. The proposed RVC model requires no transfer learning from the four training metropolitan areas to the targeted one. The main idea is to investigate the visiting circulation behaviors between origins and destinations for individuals and design a set of simple but comprehensive rules to determine the destination zone based on the given information in each trip.

This paper is organized as follows. Insights on the datasets will be discussed in detail in the Exploratory Data Analysis section (Section~\ref{sec-eda}). The proposed  Rule-based Visiting Circulation (RVC) method and its variants will be introduced in the Methodology section (Section~\ref{sec-method}). The experiments and performance evaluation of different methods compared with the baseline will be elaborated upon in the Experimental Results section (Section~\ref{sec-exp}). The conclusions will be drawn, along with discussions, in Section~\ref{sec-conclude}.

%% file: 4-ExplorativeAnalysis.tex
\section{Explorative Data Analysis}
\label{sec-eda}

\subsection{Dataset}
The dataset provided by the competition is the People Flow Dataset\footnote{\url{https://pflow.csis.u-tokyo.ac.jp/data-provision-service/about-people-flow-data/}}~\cite{pflow11,ppflow22}, which is based on person trip survey data (PT Data) collected by a country or a local authority. Each entry of the dataset represents a daily trip in the metropolitan area and contains an individual's person ID (PID), demographic features (i.e., age, gender, occupation), the origins of their trips, and the corresponding destinations. Due to privacy concerns, the locations of the origins and destinations use zones as the basic spatial units to avoid the reproduction of sensitive significant locations, such as a person's home and work locations. Nevertheless, a cartographic representation of the zones is provided in the form of shapefiles. 
Furthermore, there is also an open data-based dataset that contains features that represent each zone, including population, number of employees and offices in the secondary industry, and number of employees and offices in the tertiary industry.

The four metropolitan areas included in the training datasets are Tokyo, which contains $470,140$ travelers and $790,613$ trips, Chukyo, which contains $135,910$ travelers and $343,752$ trips, Kyushu, which contains $147,968$ travelers and $343,752$ trips, and Higa., which contains $21,968$ travelers and $34,496$ trips. The metropolitan area of the test set is Kinki, which contains $624,511$ travelers and $34,496$ trips.

\begin{figure}[!t]
  \centering
  \includegraphics[width=1.0\linewidth]{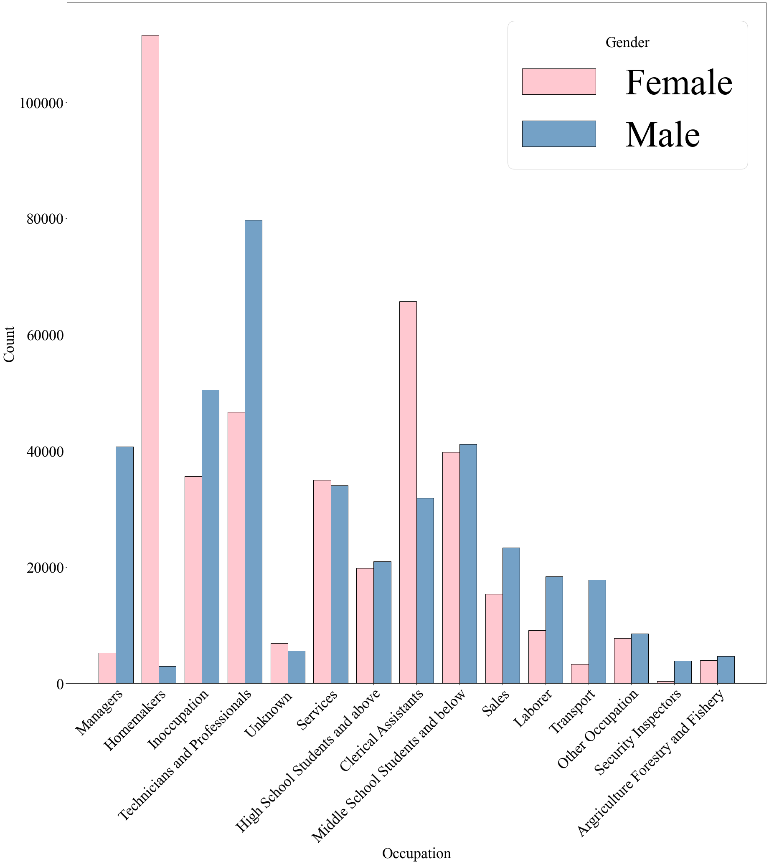}
  \caption{Occupation and Gender Histogram}
  \label{fig1}
\end{figure}

\begin{figure}[!t]
  \centering
  \includegraphics[width=1.0\linewidth]{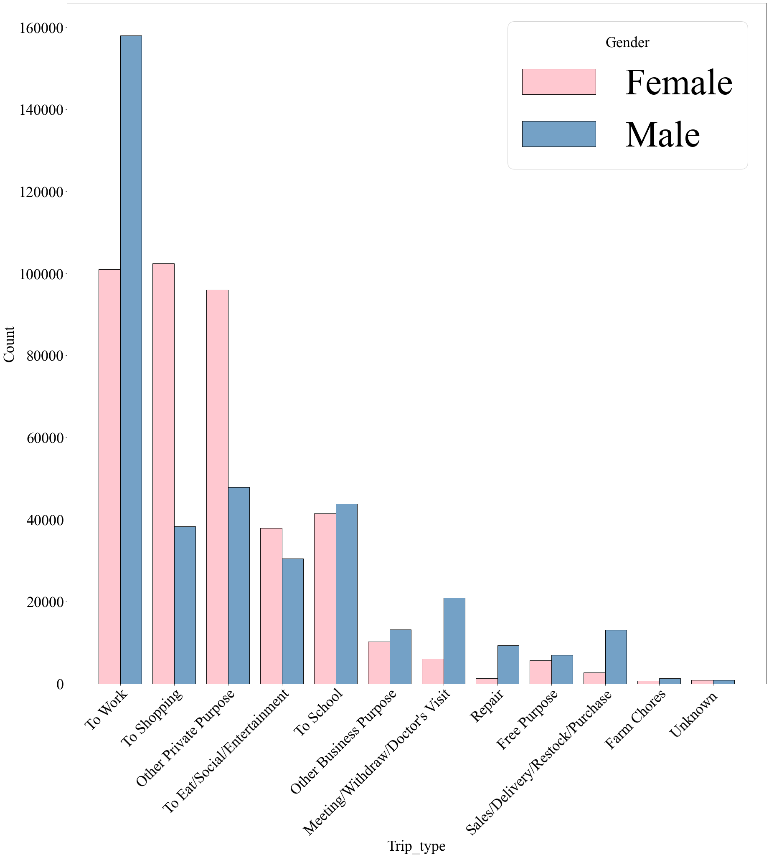}
  \caption{Trip Type and Gender Histogram}
  \label{fig2}
\end{figure}

\begin{figure}[!t]
  \centering
  \includegraphics[width=1.0\linewidth]{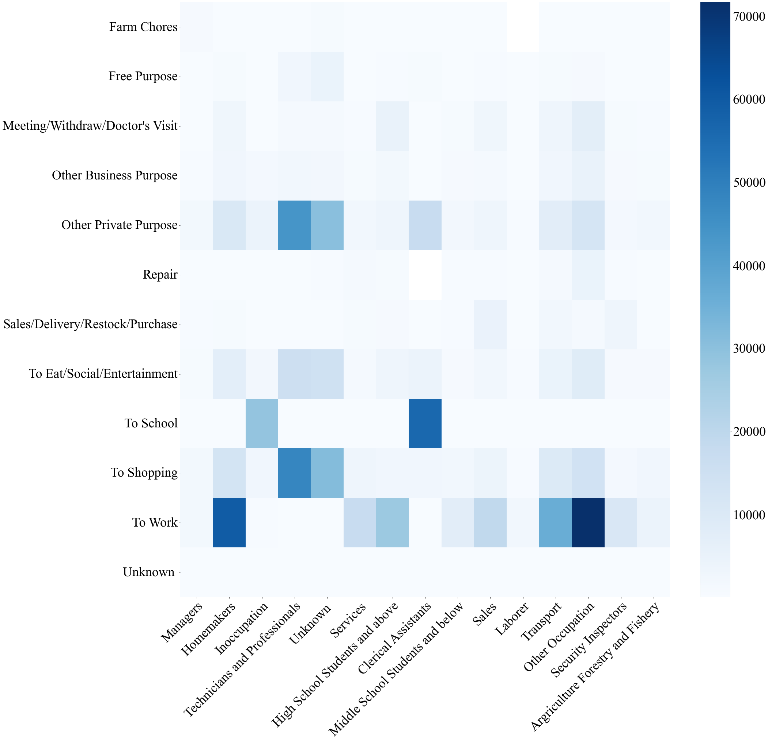}
  \caption{Trip Type and Occupation Heatmap}
  \label{fig3}
\end{figure}

\subsection{Feature Analysis}
In terms of personal attributes and trip types, there is a relatively high correlation between features. Below we use the Tokyo metropolitan area as an example to show some essential insights via explorative feature analysis.

\begin{itemize}
\item \textit{Gender and Occupation} (Figure~\ref{fig1}): For Managers, Technicians and Professionals, and Laborers, the number of men is significantly higher than women. For Homemakers Clinical Assistants and Services, it is the opposite.
\item \textit{Trip Type and Gender} (Figure~\ref{fig2}): The proportion of men that travel to work is larger than that of women, while the proportion of women that make trips to go shopping is higher than men.
\item \textit{Trip Type and Occupation} (Figure~\ref{fig3}): Students mostly make trips to school. Technicians, Professionals, and Clinical Assistants mostly make trips to work. 
\item \textit{Age and Occupation}: Occupations such as Students have a strong correlation with Age.
\end{itemize}

In this view, due to the correlation between different features, issues such as multicollinearity have to be taken into account when training prediction models. 
Furthermore, after analyzing the available features of the datasets, most of our findings proved somewhat inconclusive, as they could not make substantive improvements to the trained machine learning classification models we tried while solving the task.
We have utilized supervised learning methods, such as LightGBM and Random Forest, using features including departure time, age, gender, occupation, trip type, and origin as well as both with and without zone features. However, the prediction accuracy of these learning methods was quite low, i.e., below $0.25$, as reported in Table~\ref{tab:mainres}.

\begin{figure}[!t]
  \centering
  \includegraphics[width=1.0\linewidth]{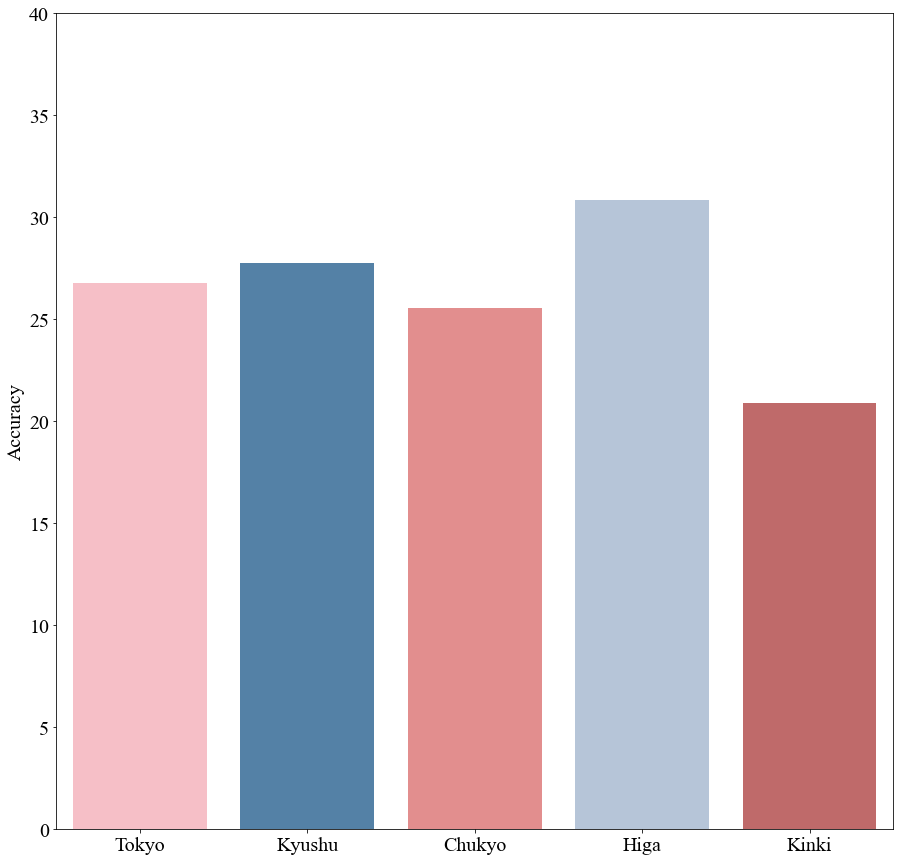}
  \caption{Accuracy of ``Destination = Origin'' in different metropolitan areas.}
  \label{fig4}
\end{figure}

\subsection{Origin and Destination Relationship}
\label{sec-odrelation}
This section serves as the basis of our proposed rule-based visiting circulation method, which will be discussed in detail in the next section. 
The relationship between Origin and Destination is interesting, stemming from the phenomenon reflected in the baseline. According to Figure~\ref{fig4}, out of over $1,000$ different destinations, trips whose Destination equals Origin take over 20\% in the test dataset, and over 25\% in average for the training datasets combined. From a distribution standpoint, this is extremely disproportionate. Nevertheless, by thinking twice logically, if an individual made a daily trip and she started out from the residence, it is certainly most likely that the last stop, i.e., the destination, will be back to where she started the trip. 

In addition, if an individual made multiple trips within a day, though the purpose of each trip may be different, the most logical speculation on that individual's destination should be the origin of her next trip~\cite{li2018route,nextloc16,origindst21}. This implies that if there are multiple trips made by the same person, the destination is most likely to be related to the origin of the trip as well as the sequence of the stops made on that trip. Such findings can be obtained in the dataset with the Origin feature and the Departure Time feature, and we will exploit these insights to devise our RVC method, whose details are described in Section~\ref{sec-method}.

%% file: 5-Model.tex
\section{The Proposed RVC Method}
\label{sec-method}
Following the analyses and findings discussed in Section~\ref{sec-odrelation}, we now present the proposed Rule-based Visiting Circulation (RVC) method to deal with the given competition task of trip destination prediction. 

In the proposed RVC, we make the assumption that every individual's daily trip is a round trip. That said, the individual starts from a location, makes one or multiple stops, then returns to the starting location~\cite{roundtrip1,roundtrip2}. Existing studies~\cite{mobi11,trp14,revisit20,checkagain,gflash22,hmt22} had also found that people have some potential to revisit locations that they visited before, and revisiting behaviors can benefit the prediction of next locations. Therefore, based on our assumption and insights from past studies, we present the rule-based visiting circulation method. If an individual makes only one trip within a day, the destination of her trip should be the origin of the trip. If multiple trips were made by the same individual, the trips would be arranged in a sequential order based on the departure time of each trip, and the destination of each trip shall be the origin of the next trip in a sequential order. The destination of the last trip should loop back to the origin of the first trip within the day, thus completing the \textit{visiting circulation}. The following sections will describe how the proposed RVC method is implemented in practice.

\subsection{Data Preprocessing}
Once we have the datasets of multiple metropolitan areas, we group all trips that have the same PID together, since trips with the same PID are those made by the same person. After the PID-based grouping of trips, we sort the trips belonging to every PID according to the Departure Time of the trip in ascending order. Such a re-ordering can ensure that trips made by every individual are arranged in a sequential order of stops they made during the day.

\begin{figure}[!t]
  \centering
  \includegraphics[width=1.0\linewidth]{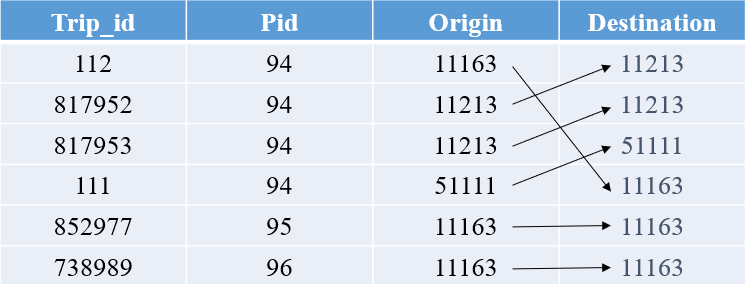}
  \caption{Elaboration on rule-based visiting circulation by assuming all trips in this table are created within a day, and all trips are sorted according to the departure time in ascending order from top to down in this table. The first four rows/trips are made by the same individual (i.e., PID $=$ 94) while the last two rows/trips are created by two different persons, i.e., PID $=$ 95, 96, respectively. The first four trips form circulation, i.e., the destination of a trip is the origin of its next trip and the destination of the last trip is the origin of the first trip. Besides, for every person who creates only one trip within a day, i.e., the last two rows, the destination is set to be the same as the origin.}
  \label{fig5}
\end{figure}

\subsection{Visiting Circulation for PIDs with Multiple Trips}
We first ensure that all trips with the same PID have been sorted in an ascending order based on the Departure Time of the trip. As illustrated by the first four rows/trips in Figure~\ref{fig5}, for trips that have the same PID, we set the prediction of the destination as the origin of the corresponding next trip that is created by the same PID in the dataset. We term such a solution as visiting circulation. 

\subsection{Visiting Circulation for PIDs with One Single Trip}
For PIDs that made only one trip in the dataset, as illustrated by the last two rows/trips in Figure~\ref{fig5}, we set the prediction of the destination as the origin of the same trip. Such a solution also follows visiting circulation, i.e., the trip is a round trip that begins and ends at the same location.

\subsection{Variants of RVC}
\label{sec-rvmvar}
The core concept of the proposed RVC is based on the assumption that for each unique individual, all daily trips are parts of a round trip. Based on the idea of visiting circulation, we can construct several variants of the RVC model where the core concept is the same, but different assumptions are made about the dataset.

\begin{enumerate}
\item \textit{Assume that the dataset is complete}: 
Trips under each PID make up a complete round trip. This is the default assumption used to implement RVC. In this assumption, since the dataset is complete, for each PID, the origin of the sequentially first trip is the starting location of the round trip and the sequentially last trip is the end of the round trip, which circulates back to the first trip. If there is only one trip in the dataset, then the trip is both the first and last trip of the round trip.

\item \textit{Assume that the dataset is incomplete}:
All trips in the dataset are parts of a round trip, but trips under the same PID may not necessarily make up a complete round trip. With such an assumption, the rules of visiting circulation can still be applied. However, the origin of the sequentially first trip may not be the starting point of the round trip. And the sequentially last trip may not be the end of the round trip, i.e., it does not necessarily have to loop back to the first trip. Possible approaches based on this assumption may include a partial implementation of visiting circulation mixed with the following two approaches.
\begin{itemize}
    \item \textit{Partial implementation of RVC mixed with other rule-based heuristics.} We can apply RVC to a part of the dataset while implementing other rule-based methods to determine or predict the destination of each trip. One possible implementation is applying RVC for individuals with more than one trip within a day and assigning the most common destination for those with only one trip based on their personal attributes. Another realization is utilizing the proposed RVC, but setting the destination to the origin for the last trip of each individual with the assumption that the individual's last trip in the dataset may not be the last trip of the circulation.
    
    \item \textit{Partial implementation of RVC mixed with supervised learning prediction models.} We can apply RVC for individuals/PIDs containing multiple trips within a day, and train a prediction model for individuals/PIDs with only one single trip. Given multiple trips of the same PID in Kinki, i.e., the training data, we apply the proposed RVC method to obtain trips with predicted destinations. These trips with RVC-based prediction outcomes will be treated as training data in the test metropolitan area, i.e., Kinki. When we are predicting the destination for individuals who create only one single trip within a day, we can train a supervised learning model, in which the prediction target is the RVC-predicted destination. The model inputs contain departure time, age, gender, trip type, occupation, origin, and zone features, and the output of the model is the destination in Kinki. By combining RVC with the learning-based method, we will be allowed to learn and generalize to Kinki.
\end{itemize}
\end{enumerate}

\subsection{Limitations of RVC}
\label{sec-limit}
Here we discuss the limitations of the proposed rule-based visiting circulation method for trip destination prediction. We also briefly deliver the potential solutions to the limitations. The limitation is two-fold, as described below.
\begin{itemize}
\item RVC is unable to detect and determine whether or not the trips created by a specific PID make up a complete round trip. This is used to be assumed in order to make the necessary adjustment for the implementation of RVC variants, as mentioned in Section~\ref{sec-rvmvar}. We believe that a significant number of trips do not form round trips, but we cannot know before the prediction made by RVC. Furthermore, round trips are not required to occur within a day, and they can happen across multiple days. Being not able to handle round trips formed across days is also a limitation of our RVC method.

\item RVC is unable to determine whether there are multiple round trips under the same PID in a given dataset. That said, RVC presumes that only one round trip within a day can happen. However, it is possible that individuals have several trips there and back centered at the origin. We think RVC can be further enhanced if we can properly divide the sequence of trips made by an individual into multiple subsequences of round trips.
\end{itemize}

To overcome these limitations, supervised learning-based methods can be investigated. For the first limitation, by utilizing the four training metropolitan areas, we can first find and label round trips, and consider the prediction of round trips as an additional task. By training on the four areas, we can infer whether trips made by an individual within a day belong to a round trip. For the second limitation, some people tend to have multiple round trips within a day. We can train a sequence segmentation model (e.g., DASSA~\cite{seqseg}) to divide an individual's trips into round trips by using the four training metropolitan areas. Then the model is applied to Kinki to find sets of trips that form round trips.

%% file: 6-Experiments.tex
\section{Experiments}
\label{sec-exp}
In this section, we conduct experiments to evaluate the performance of the proposed RVC, and have it compared with the competition baseline, the implementations of typical supervised learning-based methods, i.e., LightGBM~\cite{lightgbm}, Random Forest~\cite{randomforest}, as well as different variants of the proposed RVC. We will also evaluate the robustness of the proposed RVC, and share our final results in the leaderboard of IEEE BigData Cup 2022 Trip Destination Prediction. 

\subsection{Baseline Methods}
For evaluation purposes, the following methods are used as baselines for model comparison with the proposed RVC. It is worth noting that the performance of the methods implemented only reflects the value of these methods in this competition task. 
\begin{itemize}
\item \textit{Competition Baseline}: In this baseline, the prediction of destinations is determined by the origin of the trip. That is, for each TripID, the predicted Destination is set to the same as the Origin.
\item \textit{LightGBM}~\cite{lightgbm}: a strong gradient boosting supervised learning method with its default hyperparameter values provided by the package~\footnote{\url{https://github.com/microsoft/LightGBM}}.
\item \textit{Random Forest}~\cite{randomforest}: a typical tree-based supervised learning method with its default hyperparameter values provided by the scikit-learn package~\footnote{\url{https://scikit-learn.org/stable/modules/generated/sklearn.ensemble.RandomForestClassifier.html}}.
\end{itemize}

\subsection{Feature Preprocessing}
In terms of feature preprocessing, we simply divide the features into categorical and continuous variables. We use one-hot encoding~\cite{cerda2018similarity} for categorical variables and StandardScaler\footnote{\url{https://scikit-learn.org/stable/modules/generated/sklearn.preprocessing.StandardScaler.html}} for continuous variables. The preprocessing of each feature is specified as follows.
\begin{itemize}
\item Age: annotated by 0-5, 6-10, 11-15, ..., and 80 above, we transform the Age values to the median in the respective range, i.e., 3, 8, 13, etc. 
After that, we use StandardScaler to transform and normalize the feature values. 
\item Departure time: in each metropolitan area, all the departure time is on the same day. For example, all departure times in Tokyo are between 2008/10/01 3:00 A.M. and 2008/10/02 2:59 A.M. We then segment the time into 24 classes, in which each class represents an hour. If the departure time of one entry is 2008/10-02 7:00 A.M., we relabel it as ``7''. After the relabeling, we use one-hot encoding~\cite{cerda2018similarity} to transform the Departure Time. Thus, we are left with 24 variables representing the departure time.
\item Gender: ``1'' represents Male and ``2'' represents female, we convert value ``2'' to ``0''.
\item Occupation: We find that in different metropolitan areas, the same occupation code maps to different occupations. For example, Managers is mapped to ``9'' in Tokyo but ``1'' in Kinki. In order to solve the problem of inconsistency occupation coding in different metropolitan areas. We use ``data\_specification.csv'' and convert the occupation code to its corresponding description, such as Managers, Technicians, and Professionals, etc. After that, we find that there are ``High School Students and Above'' in Kinki, Kyushu, and Chukyo, and there are ``High School Students'' and ``Graduates and Above'' in Tokyo and Higa. Since our test set is Kinki, we combine ``High School Students'' and ``Graduates and Above'' to ``High School Students and Above'' in Tokyo and Higa. We also transform some occupations that are in other metropolitan areas but not in Kinki to an occupation that is in Kinki, ensuring that there is no occupation that is not in Kinki but in other areas. Lastly, we use one-hot encoding~\cite{cerda2018similarity} to transform the occupation coding.
\item Trip type: trip type and occupation have the same inconsistency problem with their mapping across metropolitan areas. For example, we need to combine ``Other Private Purpose,'' ``To Hospital,'' and ``Pick up'' to form the variable ``Other Private Purpose (including To Hospital and Pick up)'' in the Tokyo dataset. In addition, we also apply the same method as occupation to avoid inconsistent trip types. Lastly, we use one-hot encoding~\cite{cerda2018similarity} to transform the trip type codes.
\item Origin and destination: we add the zone features of the corresponding origin. However, the origin/destination labels of different metropolitan areas are not the same. We use zone features to calculate the cosine similarity between zones in Kinki and zones not in Kinki. Through the similarity measurement, we can map the zones of Tokyo, Chukyo, Kyushu, and Higa to the most similar zone in Kinki.
\end{itemize}

We fit the features above including age, departure time, gender, occupation, trip type, origin (most similar zone in Kinki), and origin zone features into the baseline models to predict each trip's destination zone.

\begin{table}[!t]
\centering
\caption{Performance comparison for different supervised learning methods compared with RVC.}
\label{tab:mainres}
\begin{tabular}{c|c|c}
\hline
Method & Description & Accuracy \\ \hline
Baseline & Destination equals Origin & 0.20879 \\ \hline
LightBGM & Gradient Boost & 0.13020 \\ \hline
Random Forest & Ensemble Classifier & 0.23124 \\ \hline
RVC & The default assumption of RVC & \textbf{0.43838} \\
\hline
\end{tabular}%
\end{table}

\subsection{Main Results}
The results in terms of accuracy reported in the leaderboard are exhibited in Table~\ref{tab:mainres}. We can obviously see the superiority of the performance of the proposed RVC with the default assumption mentioned in Section~\ref{sec-rvmvar} when comparing to the baseline and supervised learning methods LightGBM and Random Forest, evaluated on the Kinki metropolitan area. Nevertheless, we believe the reason for the tremendous discrepancy in performance lies in the difficulty of training a classification model based on the challenges mentioned in Section~\ref{sec-intro}, not to mention the performance of these methods compared with the competition baseline, which is clearly not ideal. Such results provide positive support for visiting circulation in predicting trip destinations even though we have no knowledge about the destinations in the test metropolitan area for model training.

\begin{table}[!t]
\centering
\caption{Performance comparison of RVC variants.}
\label{tab:varres}
\begin{tabular}{c|c}
\hline
RVC Variants & Accuracy \\ \hline
Baseline & 0.20879 \\ \hline
RVC + LightBGM & 0.37069 \\ \hline
RVC + Random Forest & 0.37798 \\ \hline
RVC + LightBGM + Zone Features & 0.37153 \\ \hline
RVC + Random Forest + Zone Features & 0.41727 \\ \hline
RVC (Default Assumption) & 0.43838 \\ \hline
\end{tabular}%
\end{table}

\subsection{Results on RVC Variants}
Due to the tremendous discrepancy seen in Table~\ref{tab:mainres}, in the second round of our experiments, we aim to see how different variants of the proposed RVC (described in Section~\ref{sec-rvmvar}) yield various performances. 
For instance, RVC $+$ LightGBM means that we first apply RVC to generate the predicted destinations for multiple trips created by the same PID, and consider them as training data for LightGBM to learn and construct the supervised destination predictor based on the feature preprocessing methods mentioned previously. Then the trained LightGBM model is utilized to predict the destinations of the remaining trips in the Kinki area.
The results of the performance comparison can be seen in Table~\ref{tab:varres}. once RVC is partially integrated into the supervised learning models, their accuracy scores improve drastically when evaluating on the Kinki metropolitan area. Take Random Forest as an example, its accuracy increases from $0.23124$ in Table~\ref{tab:mainres} to $0.37798$ in Table~\ref{tab:varres}. Such results prove that applying RVC first to the test metropolitan area to obtain labeled destination data for training can significantly improve the generalization capability for supervised learning methods. In addition, we can also see that once the zone features have been added, a notable increase in performance can also be seen.

\begin{table}[!t]
\centering
\caption{Statistics for PIDs with multiple trips (i.e., PID trip count \textgreater 1) on various metropolitan areas.}
\label{tab:areastat}
\begin{tabular}{c|ccccc}
\hline
 & \multicolumn{5}{c}{PID trip counts \textgreater 1} \\ \hline
 & \multicolumn{1}{c|}{Tokyo} & \multicolumn{1}{c|}{Chukyo} & \multicolumn{1}{c|}{Kyushu} & \multicolumn{1}{c|}{Higa.} & Kinki \\ \hline
Count & \multicolumn{1}{c|}{508,341} & \multicolumn{1}{c|}{113,820} & \multicolumn{1}{c|}{313,821} & \multicolumn{1}{c|}{20,095} & 548,834 \\ \hline
Proportion & \multicolumn{1}{c|}{0.6430} & \multicolumn{1}{c|}{0.5540} & \multicolumn{1}{c|}{0.9129} & \multicolumn{1}{c|}{0.5825} & 0.5673 \\ \hline
\end{tabular}%
\end{table}

\begin{table}[!t]
\centering
\caption{Performance comparison between baseline and RVC across metropolitan areas.}
\label{tab:areares}
\begin{tabular}{c|ccccc}
\hline
 & \multicolumn{5}{c}{Accuracy} \\ \hline
 & \multicolumn{1}{c|}{Tokyo} & \multicolumn{1}{c|}{Chukyo} & \multicolumn{1}{c|}{Kyushu} & \multicolumn{1}{c|}{Higa.} & Kinki \\ \hline
Baseline & \multicolumn{1}{c|}{0.26754} & \multicolumn{1}{c|}{0.25541} & \multicolumn{1}{c|}{0.27737} & \multicolumn{1}{c|}{0.30838} & 0.20879 \\ \hline
RVC & \multicolumn{1}{c|}{0.51291} & \multicolumn{1}{c|}{0.43620} & \multicolumn{1}{c|}{0.69231} & \multicolumn{1}{c|}{0.50195} & 0.43838 \\ \hline
\end{tabular}%
\end{table}

\begin{table*}[!t]
\centering
\caption{Results on ablation study for RVC.}
\label{tab:ablres}
\begin{tabular}{c|c|c}
\hline
Method & Description & Accuracy \\ \hline
Baseline & Destination = Origin for all trips & 0.20879 \\ \hline
Partial RVC & only on trips where the PID trip counts \textgreater 1, others set to 0 & 0.28156 \\ \hline
Full RVC & RVC on trips where the PID trip counts \textgreater 1 and Destination = Origin for the others & 0.43838 \\ \hline
\end{tabular}%
\end{table*}

\begin{figure*}[!t]
  \centering
  \includegraphics[width=0.8\textwidth]{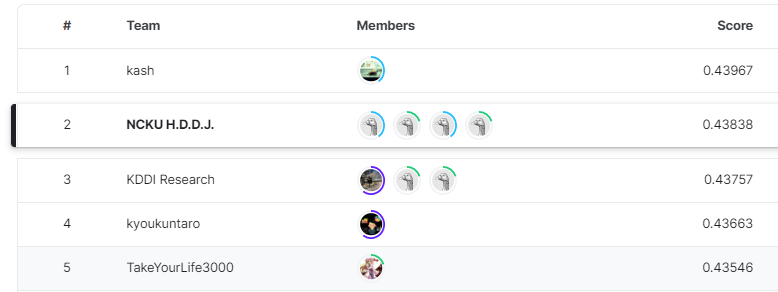}
  \caption{The final result of the proposed RVC method (Team ``NCKU H.D.D.J.'') in the competition leaderboard.}
  \label{fig6}
\end{figure*}

\subsection{Robustness Evaluation}
Since the proposed RVC is a rule-based method, it should be relatively robust due to the fact that it does not rely on any prior information or training data. To examine the robustness of the proposed RVC, we conduct the experiments using the datasets in the four training metropolitan areas. We wonder whether RVC can again lead to a promising performance as good as the accuracy in Kinki.
By removing all of the destinations for each trip of the training metropolitan area, we can construct the artificial testing data whose setting is the same as that in Kinki. Since the main idea of visiting circulation is established upon individuals with multiple trips in a day, i.e., a round-trip usually consists of multiple stops before returning to the starting location, here we focus on examining the performance of RVC for PIDs with multiple trips (i.e., PID trip count $>$ 1). The statistics of the compiled artificial test datasets for the four metropolitan areas are shown in Table~\ref{tab:areastat}. 
We can see the proportion values vary across metropolitan areas, in which the Kyushu area has the highest proportion of PIDs with multiple trips.
We think that the higher the proportion of PID trip counts $>$ 1, the more likely a complete round trip is included for each PID. And more complete round trips can better fit the idea of visiting circulation and have a higher possibility to generate 
higher accuracy scores for the proposed RVC in trip destination prediction. 

We apply RVC to every artificial testing data and report the accuracy scores in Table~\ref{tab:areares}.
It can be apparently observed that our insight on the relationship between a higher proportion of multi-trip PIDs and better performance does hold by corresponding metropolitan areas in Table~\ref{tab:areastat} with those in Table~\ref{tab:areares}. That said, the performance of destination prediction in the Kyushu area leads to the highest accuracy score.
In this view, we can see that aside from the correlation between the proportion of multi-trip PIDs and prediction performance, the proposed RVC also proves to be fairly robust as it yields pretty good results for all metropolitan areas.

\subsection{Ablation Study}
We further conduct an ablation study to investigate whether the two ideas in visiting circulation, i.e., for multi-trip PIDs and single-trip PIDs within a day, all contribute to the performance. We perform ``Partial RVC'' by applying RVC to only PIDs with multiple trips in Kinki metropolitan area and randomly predicting the destinations for single-trip PIDs, and compare it with the Full RVC (i.e., applying RVC to both multi-trip and single-trip PIDs). The results of the ablation study are exhibited in Table~\ref{tab:ablres}. 
We can see that the accuracy of Partial RVC is not satisfying, and not as good as Full RVC. Such results prove the usefulness of applying RVC to single-trip PIDs and verify the effectiveness of visiting circulation.

\subsection{Final Leaderboard Results}
The snapshot of final competition results using the proposed RVC in the leaderboard\footnote{\url{https://www.kaggle.com/competitions/ieee-bigdata-cup-2022-destination-prediction/leaderboard}} is exhibited in Figure~\ref{fig6}. Our team name is ``NCKU H.D.D.J.'', and the final accuracy of RVC is $0.43838$, which is ranked at the second place in both the public and private leaderboards since the private leaderboard is calculated over the same rows as the public one.

%% file: 7-Conclusion.tex
\section{Conclusions}
\label{sec-conclude}
In this paper, we propose a simple Rule-based Visiting Circulation (RVC) approach to solve the IEEE BigData Cup 2022: Trip Destination Prediction competition and scored an accuracy of $0.43838$, which stands at second place on the final leaderboard of the competition website on Kaggle. 

Although there are a lot of data and resources available in the datasets provided as well as online, the proposed RVC, though simple, significantly outperforms typical supervised learning-based classification or prediction methods. Furthermore, the only methods that could put up a fight against the proposed RVC are its variants, i.e., the mix of RVC and Random Forest as well as the mix of RVC and LightGBM. With such promising performance, it is safe to say that the proposed RVC is a competitive approach in this Trip Destination prediction challenge, while at the same time proving that supervised learning techniques, though useful on many occasions, do not always yield the best results if used blindly. In addition, it is also worthwhile noticing that to have the generalization capability for supervised learning methods, one should carefully create supervision signal (i.e., labeled information) in the testing domain. In the competition task of trip destination prediction, RVC can serve as a proper way to bring destination labels for model training of supervised methods. Experimental results also prove that RVC is a simple but quite effective rule-based method with great robustness and promising generalization capability. 

Last, despite the success we achieved in this competition with RVC, there is still much that can be improved. e.g., regarding the competition, we do not know how RVC will face against models that successfully integrate geographic and spatial features into their models. Also, the limitations of RVC, as mentioned in Section~\ref{sec-limit} are yet to be solved.